\documentclass[conference]{IEEEtran}

\usepackage{amsmath}
\usepackage{mathtools}
\usepackage [english]{babel}
\usepackage [autostyle, english = american]{csquotes}
\usepackage{multirow}

\MakeOuterQuote{"}

\usepackage[flushleft]{threeparttable}

\DeclareMathOperator*{\argmin}{arg\,min}

\makeatletter
\newcommand{\mypm}{\mathbin{\mathpalette\@mypm\relax}}
\newcommand{\@mypm}[2]{\ooalign{%
  \raisebox{.1\height}{$#1+$}\cr
  \smash{\raisebox{-.6\height}{$#1-$}}\cr}}
\makeatother

\begin{document}

\title{An Unsupervised Approach for Overlapping Cervical Cell Cytoplasm Segmentation}

\author{
    \IEEEauthorblockN{Pranav Kumar\IEEEauthorrefmark{1}, S L Happy\IEEEauthorrefmark{2}, Swarnadip Chatterjee\IEEEauthorrefmark{3}, Debdoot Sheet\IEEEauthorrefmark{2}, Aurobinda Routray\IEEEauthorrefmark{2}}
    \IEEEauthorblockA{\IEEEauthorrefmark{1}Computer Science and Engineering, National Institute of Technology, Tiruchirappalli, India}
    \IEEEauthorblockA{\IEEEauthorrefmark{2}Department of Electrical Engineering, Indian Institute of Technology, Kharagpur, India}
    \IEEEauthorblockA{\IEEEauthorrefmark{3}Advanced Technology Development Center, Indian Institute of Technology, Kharagpur, India}
}

\maketitle

\begin{abstract}
The poor contrast and the overlapping of cervical cell cytoplasm are the major issues in the accurate segmentation of cervical cell cytoplasm. This paper presents an automated unsupervised cytoplasm segmentation approach which can effectively find the cytoplasm boundaries in overlapping cells. The proposed approach first segments the cell clumps from the cervical smear image and detects the nuclei in each cell clump. A modified Otsu method with prior class probability is proposed for accurate segmentation of nuclei from the cell clumps. 
Using distance regularized level set evolution, the contour around each nucleus is evolved until it reaches the cytoplasm boundaries. Promising results were obtained by experimenting on ISBI 2015 challenge dataset.
\end{abstract}


\IEEEpeerreviewmaketitle

\section{Introduction}
The Papanicolaou test \cite{papanicolaou1942new} (also known as Pap test) is a cervical screening procedure for detecting cervical cancer. In Pap test, the cells from the cervix are gently excoriated and then examined under a microscope to detect the presence of any abnormalities. These abnormal results indicate the presence of precancerous changes, yielding examinations and probable preventive treatments. This test still exists as a crucial modality in distinguishing the precursor lesions for cervical cancer. For automatic analysis of the Pap smear, the cervical cells are to be detected and segmented efficiently. However, the presence of blood, inflammatory cells, mucus and other debris makes the detection process erroneous. Moreover, the cytoplasm of cervical cells has poor contrast and irregular shape. The relative difference between light and dark areas of the smear is amiss. 

The overlapping of the cervical cells is another factor that aggravates the inter- and intra-observer variability and subsequently produces a large variance in false negative results. This forges their segmentation process as another major issue. Contemporary systems can segment the nucleus and cytoplasm in non-overlapping cervical cells \cite{li2012cytoplasm}, and also segment overlapping nuclei \cite{plissiti2012overlapping} and the cell clumps. But the pertinent methods that are used for the complete segmentation of those overlapping cells are not robust enough for clinical routines. The upper layer of cells(in the several layers of the overlapping cervical cells) partially obstructs the lower layer of cells, as a result, these methods are severely complicated and made ineffective. 

Automatic thresholding, morphological operations, and active contour models are primary approaches to the segmentation of the cervical cells. Wu \textit{et al.}\cite{wu1998optimal} detected the boundary of the nuclei of overlapping cervical cells by solving an optimal thresholding problem. Morphological analysis based approach was reported in \cite{plissiti2011automated} to detect the overlapping nuclei in the cell clumps. Morphological watersheds \cite{gencctav2012unsupervised} has also been used for the nuclei and cytoplasm separation. Other techniques \cite{yang2008edge} has proven to be efficient in segmenting the nucleus as well as the cytoplasm of isolated cells.
Active contour based methods are also quite popular in segmenting the boundaries of the cell cytoplasm \cite{li2012cytoplasm}. Other mechanisms which has also been employed are based on template fitting \cite{garrido2000applying}, edge detectors \cite{lin2009detection}, level set functions \cite{lu2013automated} and region growing with moving K-means \cite{isa2005automated} etc. 

Recently, a few researchers used the images of different focal planes to improve the accuracy of cell segmentation. For instance, Phoulady \textit{et al.} \cite{phouladyapproach} detected the nuclei and cell clumps from smear images using iterative threshold method and two component Gaussian mixture model respectively. Further, the cytoplasm boundary detection was carried out on images of different focal planes based on the edge strength of grid squares. In \cite{ushizima2014segmentation}, the nuclei were first detected from low contrast cell clump images and the cytoplasm boundary was segmented by using Voronoi diagram of detected nuclei. Similar methods were carried out in \cite{ramalho2015cell}, where the superpixel combined to Voronoi Diagrams was used as an initial estimate of cytoplasm boundary which was further refined by edge maps and morphological operations.  

This paper proposes an unsupervised method of cell segmentation which is applicable for an automatic and accurate segmentation of the overlapping cervical cells. With the prior knowledge of class probability (here nucleus), we modified the Otsu's threshold method for accurate segmentation of nuclei from the cell clumps. Further, the cytoplasm is segmented from the cell clumps using the distance regularized level set evolution which produces significant results for an unsupervised method.

\section{Proposed Approach} \label{proposed_approach}
\begin{figure*}[!t]
\centering
\mbox{\includegraphics[width=0.65\textwidth]{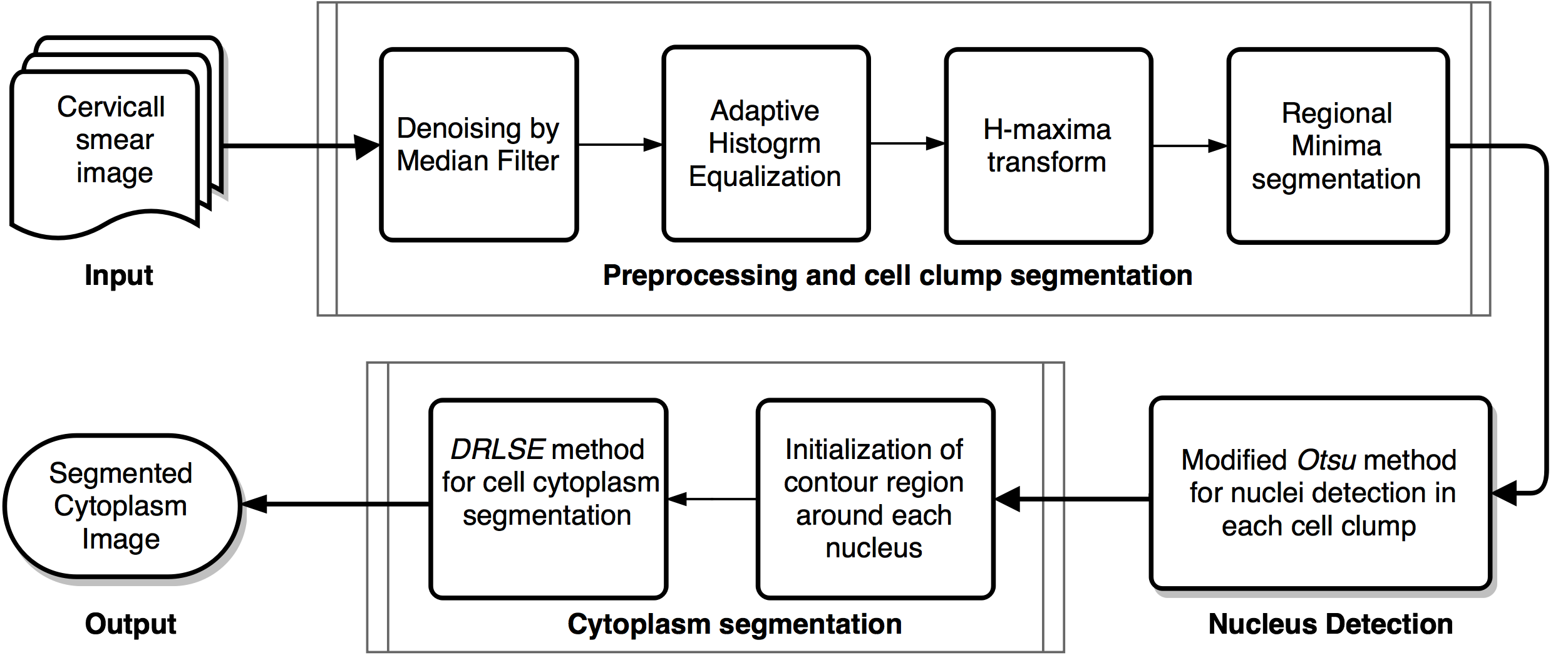}}
\caption{Overview of our approach}
\label{proposedframework}
\end{figure*}
Lu et al. \cite{lu2013automated} proposed an overlapping cell segmentation method in which cell clump segmentation is carried out first followed by nuclei detection. They used one nucleus per cell assumption to further segment the overlapping cells minimizing an energy function. A similar method is adopted here. 
The proposed unsupervised approach implements three stages to successfully segment the overlapping cervical cell cytoplasm which is shown in Fig. \ref{proposedframework}. First, the cell clumps were segmented from the Pap image. Second, each cell clump was processed individually to segment the nucleus and obtain the nucleus count in that cell clump. Third, depending upon the nucleus position and nucleus count, distance regularized level set model \cite{li2010distance} was used in each cell clump for cytoplasm segmentation. The number of cells in a specimen was assumed to be equal to the count of nuclei, thereby segmenting the cell cytoplasm from the previously segmented cell clump.

\subsection{Cell clump segmentation} \label{cell_clump_segmentaion}
The Pap smear image was pre-processed to remove noise and unwanted small grain-like structures. First, a median filter of size $5\times5$ was applied followed by adaptive histogram equalization. The median filter is a non-linear digital filtering technique, which removes the unwanted noise, such as small debris, tiny blood cells in a pap smear image. The adaptive histogram equalization improves the contrast of the image, thereby improving the cell clump segmentation accuracy.

\begin{figure}[!b]
\centering
\includegraphics[width=0.5\textwidth]{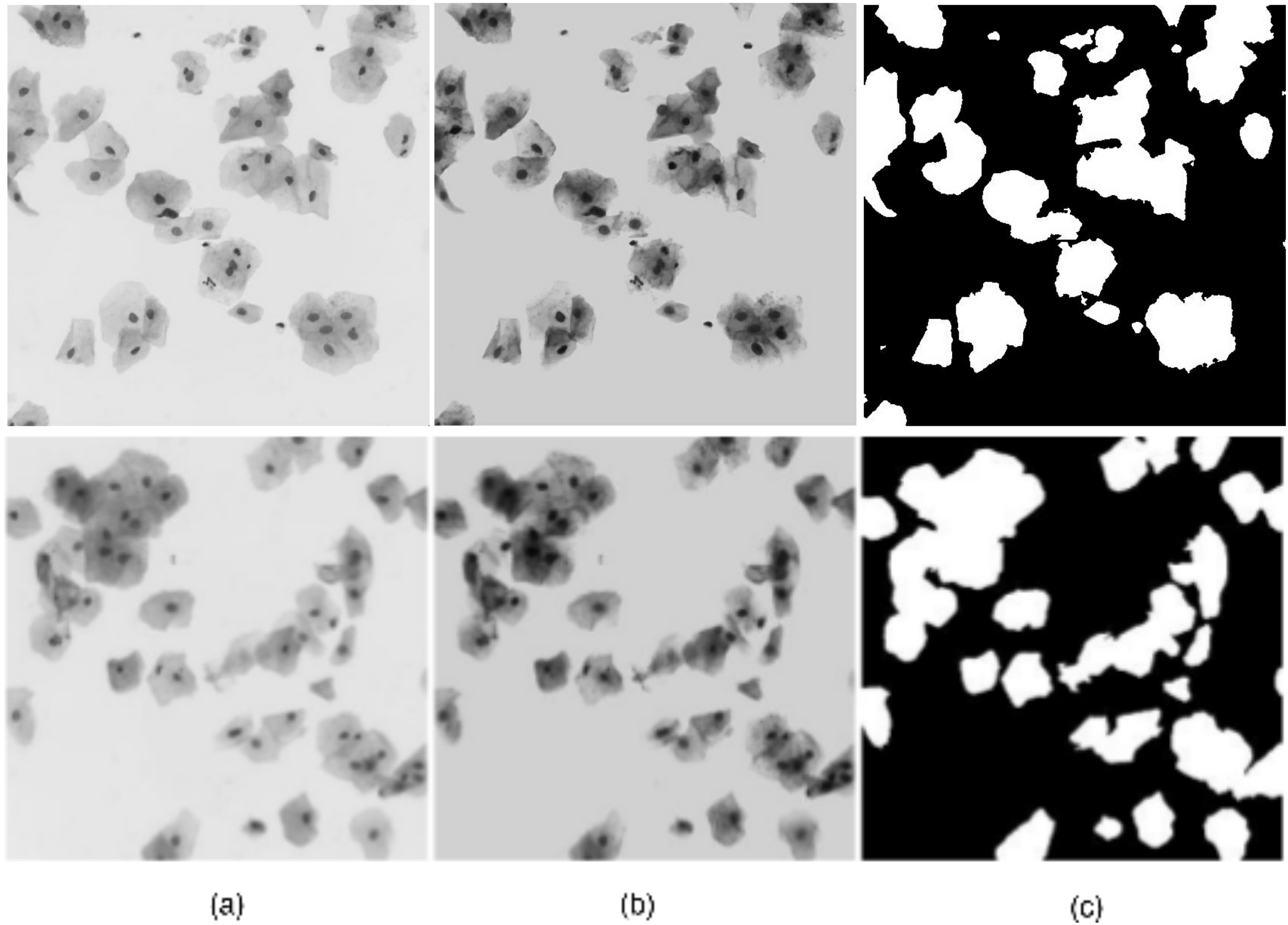}
\caption{Results of scene segmentation. (a) The cervical smear images, (b) prepossessed images for segmentation, (c) cell clump segmentation.}
\label{clump_seg}
\end{figure}

Further, H-maxima transform was applied to suppress all the maxima in the intensity image and then the binary image was obtained using the regional minima of the image. The regional minima algorithm finds the connected components in an image with a constant intensity value with the boundary pixels having a higher value than the inner area. Regional minima segments the cell clump since the cell cytoplasm in our image is of low intensity than the background. The unwanted small areas in the binary image were removed by removing all connected components having less than an area threshold which was approximately equal to the area of smallest cytoplasm present in an image. Each connected component was considered as a cell clump.

Fig. \ref{clump_seg}(a) shows the initial cervical images of two specimens. These were processed using a median filter and adaptive histogram equalization, to bring about the pre-processed images as in Fig. \ref{clump_seg}(b). These were further converted to binary images by utilizing the regional minima as a level of discrimination between black and white, aiding us to enact the cell clump segmentation as in Fig. \ref{clump_seg}(c).

\subsection{Modified Otsu method with class prior probability}

The Otsu's method \cite{otsu1975threshold} is one of the most prominent techniques to perform automated clustering-based image thresholding. This method operates directly on the gray-level histogram, therefore, it is computationally efficient \cite{otsu1975threshold} once the histogram is obtained. We propose a modified Otsu threshold approach for nucleus segmentation, which seems to be a  challenging task since the contrast of the overlapping cytoplasm is not uniform. 

\begin{figure}[!t]
\centering
\includegraphics[width=0.4\textwidth]{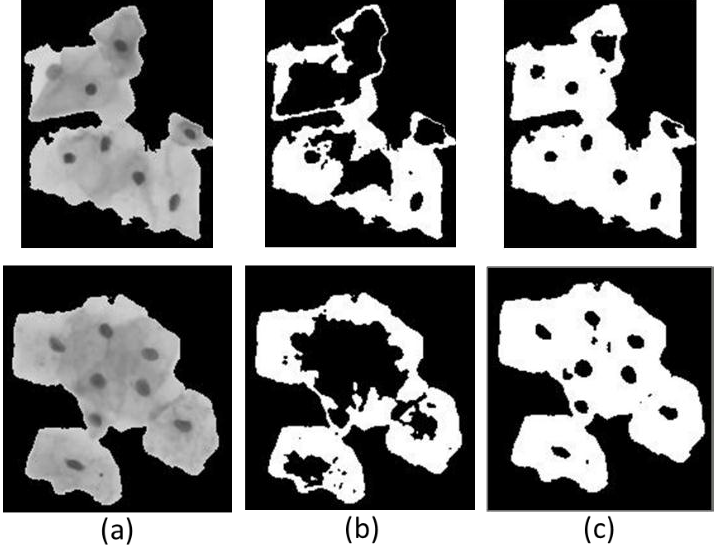}
\caption{Nucleus segmentation, (a) cell clump, (b) Otsu’s method, (c) modified Otsu with prior class probabilities (here =0.05).}
\label{D2}
\end{figure}

Otsu's method assumes bi-modal distribution of gray-level intensities and the threshold can be obtained by minimizing the within-class variances \cite{xu2011characteristic}, given by
\begin{equation} \label{eqn1}
T = \argmin_{1<T<L_{max}} \{\sigma_w^2(T) \}
\end{equation}

The within-class variance ($\sigma_w^2$) can be written in terms of class variance ($\sigma_i$) and class probabilities ($\omega_i$) as

\begin{equation}
\sigma_w^2(t) = \omega_1(t)\sigma_1^2(t) + \omega_2(t)\sigma_2^2(t)
\end{equation}

Here, $T$ is the threshold separating the classes. Given the histogram ($h$) of an image, we can find the probability of each pixel $p(i) = \frac{h(i)}{Total Pixels}$  ,$i$ = 1,$\ldots$,$L_{max}$ ($L_{max}$ = maximum pixel label). The class probabilities can be calculated from the image histogram as
\begin{equation}
\omega_1(t) = \sum_{i=1}^t p(i) \quad \textrm{and} \quad 
\omega_2(t) = \sum_{i=t+1}^{L_{max}} p(i)
\end{equation}

Xu et al. \cite{xu2011characteristic} reported that Otsu threshold deviates from the intersection point toward the center of the class having higher variance. Hence, it is applicable for segmentation if both classes have comparable within-class variances. However, nucleus segmentation from the cell clump falls under the category of unequal class probabilities. The proportion of nucleus pixels in a cell clump is far less than the portion of cytoplasm. Therefore, we modified the distribution of the pixels based on the prior class probabilities to segment nucleus accurately.

The proposed prior probability based modified Otsu method uses the prior knowledge of rough class probability to find suitable threshold when the class probability of one class is very less compared to the other. Given the prior class probability of the class containing dark pixels, $\alpha$ , we determine the maximum value of $l$ such that $\sum_{i=1}^l p(i) < \alpha$. The pixel distributions are further modified as,
\begin{equation}
p'(i) = \begin{cases}
 p(i)*(1-\alpha)       & \quad \text{if } i < l\\
    p(i)*\alpha  & \quad \text{otherwise }\\
\end{cases}
\end{equation}
\begin{equation}
p_{new}(i) = \frac{p'(i)}{\sum_i p'(i)}
\end{equation}

If $\alpha$ is small, then the pixel probabilities up to $l$ are given a higher weight than the rest and vice versa. The threshold is further determined using $p_{new}(i)$ as the pixel probabilities. Thus, the class distribution is modified to drag the threshold to the intersection point. Fig. \ref{D2} shows the differences in nucleus segmentation results using the normal Otsu's method and the modified Otsu with a prior probability of nucleus region in a cell clump as $0.05$.

\subsection{Cytoplasm segmentation}
Once the number of nuclei in each cell clump was determined, the cytoplasms of the corresponding cells were determined by using level set evolution. Assuming that the nuclei do not overlap, the number of cells in an image is equal to the number of nuclei obtained. If the number of the nucleus in the connected region was one, then the connected region was considered as one cell. In the case of multiple nuclei in a clump, we initialized the cell region with a disc (contour) surrounding the nucleus which further evolves toward the cytoplasm edges and the corresponding cytoplasm was segmented.

\begin{figure}[!h]
\centering
\includegraphics[width=0.5\textwidth]{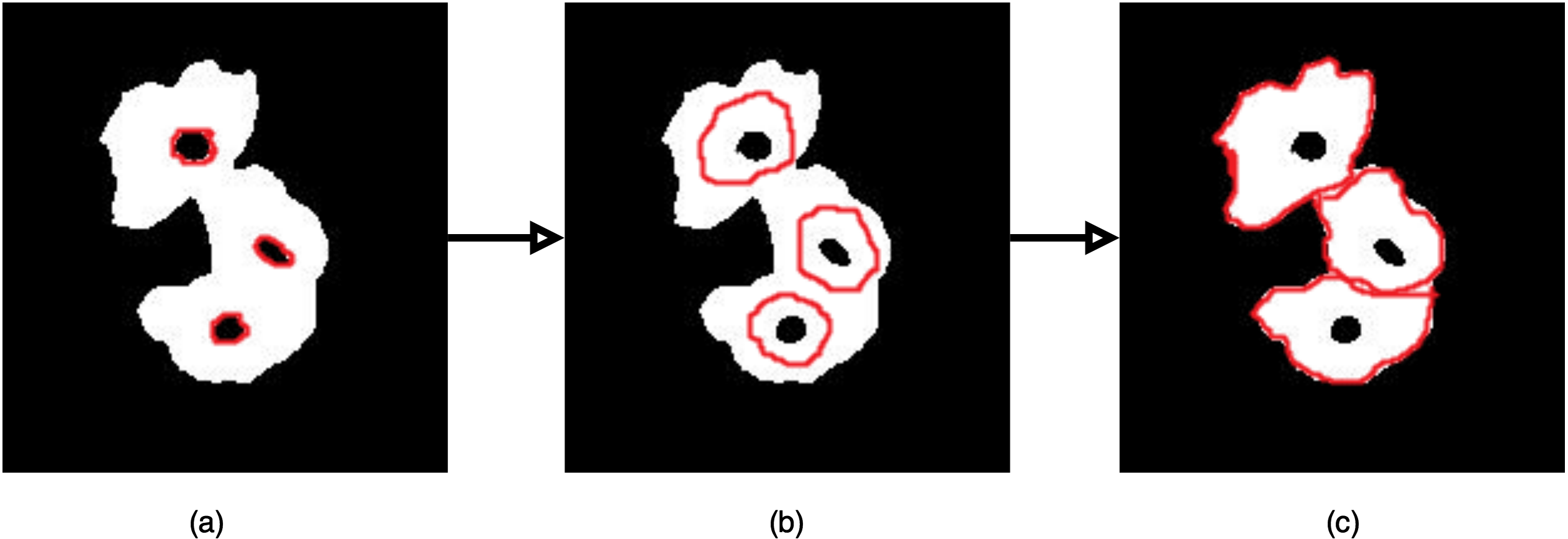}
\caption{Cytoplasm segmentation through contour evolution of DRLSE (a) initial contour, (b) inter-mediate step, (c) stopped by the edges (final cell cytoplasm)}
\label{cyt_seg}
\end{figure}

The cytoplasm segmentation process is displayed in Fig. \ref{cyt_seg}. We used the distance regularized level set evolution (DRLSE) model \cite{li2010distance} for cytoplasm segmentation, by which a contour is propagated outward from the position of the cell nucleus as shown in Fig. \ref{cyt_seg}(a). And it is stopped once it reaches the edges as in Fig. \ref{cyt_seg}(c). The major advantages of DLRSE model over traditional level set methods are that it allows more flexible and efficient initialization, and ensures accuracy with a stable level set evolution. Moreover, it is computationally efficient and allows the use of large time steps to considerably speed up the curve evolution.

\section{Experimental Results}

The proposed method was evaluated using the test set provided in ISBI 2015 challenge. The dataset provided in "The Second Overlapping Cervical Cytology Image Segmentation Challenge" contains multi-layer cytology volumes each consisting of multi-focal images acquired from the same specimen. A stack of 20 images, each of size 1024x1024 pixels, at different focal planes are provided for each specimen image and they contain solitary or overlapping cervical cells with different degrees of overlapping. The contrast and the texture of the cells are also not consistent. The test set consisted of depth images of 9 specimens with ground truth of cell cytoplasm. 

We first obtained the extended depth of field (EDF) image \cite{pertuz2013generation} for each specimen by combining their respective stack of 20 multi-focal images. Further pre-processing on EDF image was carried out for cell clump segmentation followed by nucleus detection and cytoplasm segmentation as discussed in section \ref{proposed_approach}.

The obtained cytoplasm segmentation results on testing sets are compared with the results attained by the other state-of-art techniques in the Table \ref{table1}. The results in Table \ref{table1} are presented in terms of Dice Coefficient, False Negative (object level), True Positive (pixel level) and False Positive (pixel level). The quantitative performance of individual cell cytoplasm segmentation is assessed using the average Dice Coefficient (DC). DC, ranging from 0 to 1, is the similarity measure used for comparing two samples, measured as
\begin{equation}
DC = \frac{2|X \cap Y|}{|X|+|Y|}
\end{equation} 
where $X$ and $Y$ are the two samples. The segmentation is considered to be good if $DC>0.7$. The True Positive Rate (TPR) or the Sensitivity measures the proportion of the correctly segmented pixels to total segmented pixels. The False Positive Rate (FPR) is the proportion of segmented pixels which are actually negative, to the total segmented pixels. Here the object-based false negative rate (FNO) is obtained as the proportion of cells having a $DC \le 0.7$. The table compares four characteristics of our evaluation results with those of their results, where better evaluation occurs with higher DC and TPR values but with lower FPR and FNO values.

\begin{table}[t]
\centering
\caption{Quantitative Comparison of our results with other competitors on the Test sets}
\label{table1}
\begin{threeparttable}
\renewcommand{\arraystretch}{1.5}
\setlength\tabcolsep{4.5pt}
\begin{tabular}{|c|c|c|c|}
\hline

\textbf{Results}  & \textbf{Our Results} & \textbf{Phoulady et al. \cite{phouladyapproach}}  & \textbf{Ushizima et al. \cite{ramalho2015cell}}  \\
\hline

\textbf{DC} & $0.852\mypm0.076$                   & $0.831\mypm0.079$ & $\mathbf{0.856\mypm0.078}$\\
\hline

\textbf{TPR} & $0.885\mypm0.101$                   & $\mathbf{0.927\mypm0.098}$ & $0.899\mypm0.113$\\
\hline

\textbf{FPR} & $\mathbf{0.0015\mypm0.001}$                   & $0.003\mypm0.002$ & $0.002\mypm0.001$\\
\hline

\textbf{FNO} & $\mathbf{0.361\mypm0.158}$                   & $0.408\mypm0.163$ & $0.501\mypm0.180$\\
\hline

\end{tabular}
\begin{tablenotes}
\small
\item DC: Dice Coefficient; TPR: True positive rate (pixel-level); FPR: False positive rate (pixel-level); FNO: False negative (object-level);
\end{tablenotes}
\end{threeparttable}
\end{table}

As can be seen in Table \ref{table1}, The proposed method produced comparable results to the state-of-art techniques. The DC obtained in the proposed method is 0.852 whereas the winner of the competition \cite{ramalho2015cell} got the DC value of 0.856. So, the DC and TPR values of the proposed method is close to the best performed methods, while the performance of the proposed approach is better for FPR and FNR values (the lower their values, the better they are). We obtained an approximate FNO value of 0.361 which is better than the other methods.

The proposed method is completely unsupervised. Moreover, it does not include the information from different focal planes of Pap smear. Still, the segmentation results obtained through this is quite significant in comparison to the state-of-art techniques which uses the supervised methods or use the information from different focal planes.

\section{Conclusion}
This paper proposes a robust and automated unsupervised cytoplasm segmentation method that offers significant performance even with a high degree of overlapping. A modified Otsu method with prior class probability is proposed which segments the nuclei from the cell clumps accurately. Further, the contours are evolved around the nuclei to the cell boundaries. This technique can be directly applied in any conventional stained cervical smear image. 

The cytoplasm segmentation can further be improved by including the edge information of cytoplasm boundary at different focal planes. Moreover, the combination of other techniques like Voronoi diagram, watershed transform, active contours etc. along with this may improve the accuracy. 

\bibliographystyle{IEEEtran}
\bibliography{ref}

\end{document}